\documentclass{article}
\usepackage{arxiv}

\usepackage[utf8]{inputenc} 
\usepackage[T1]{fontenc}    
\usepackage{hyperref}       
\usepackage{url}            
\usepackage{booktabs}       
\usepackage{amsfonts}       
\usepackage{nicefrac}       
\usepackage{microtype}      
\usepackage{lipsum}
\usepackage{graphicx}
\usepackage{wrapfig}
\usepackage{tabularx}
\usepackage{multirow}
\usepackage{enumitem}
\usepackage{amsmath,amsfonts,amssymb,amsthm,bm}
\usepackage{color}
\usepackage{subcaption}
\usepackage{amsmath}
\usepackage{footnotebackref}

\graphicspath{ {./ } }

\title{ZIGNeRF: Zero-shot 3D Scene Representation  \\ with Invertible Generative Neural Radiance Fields}

\author{\quad Kanghyeok Ko \quad \qquad \qquad Minhyeok Lee\thanks{Corresponding author.}\\ 
    \texttt{dogworld12@cau.ac.kr} \qquad \; \texttt{mlee@cau.ac.kr} \\
    \\School of Electrical and Electronics Engineering, Chung-Ang University\\}

\begin{document}
\maketitle

\begin{abstract}
Generative Neural Radiance Fields (NeRFs) have demonstrated remarkable proficiency in synthesizing multi-view images by learning the distribution of a set of unposed images. Despite the aptitude of existing generative NeRFs in generating 3D-consistent high-quality random samples within data distribution, the creation of a 3D representation of a singular input image remains a formidable challenge. In this manuscript, we introduce ZIGNeRF, an innovative model that executes zero-shot Generative Adversarial Network (GAN) inversion for the generation of multi-view images from a single out-of-domain image. The model is underpinned by a novel inverter that maps out-of-domain images into the latent code of the generator manifold. Notably, ZIGNeRF is capable of disentangling the object from the background and executing 3D operations such as 360-degree rotation or depth and horizontal translation. The efficacy of our model is validated using multiple real-image datasets: Cats, AFHQ, CelebA, CelebA-HQ, and CompCars.
\end{abstract}

\begin{figure}[t]
    \centerline{\includegraphics[width=\columnwidth]{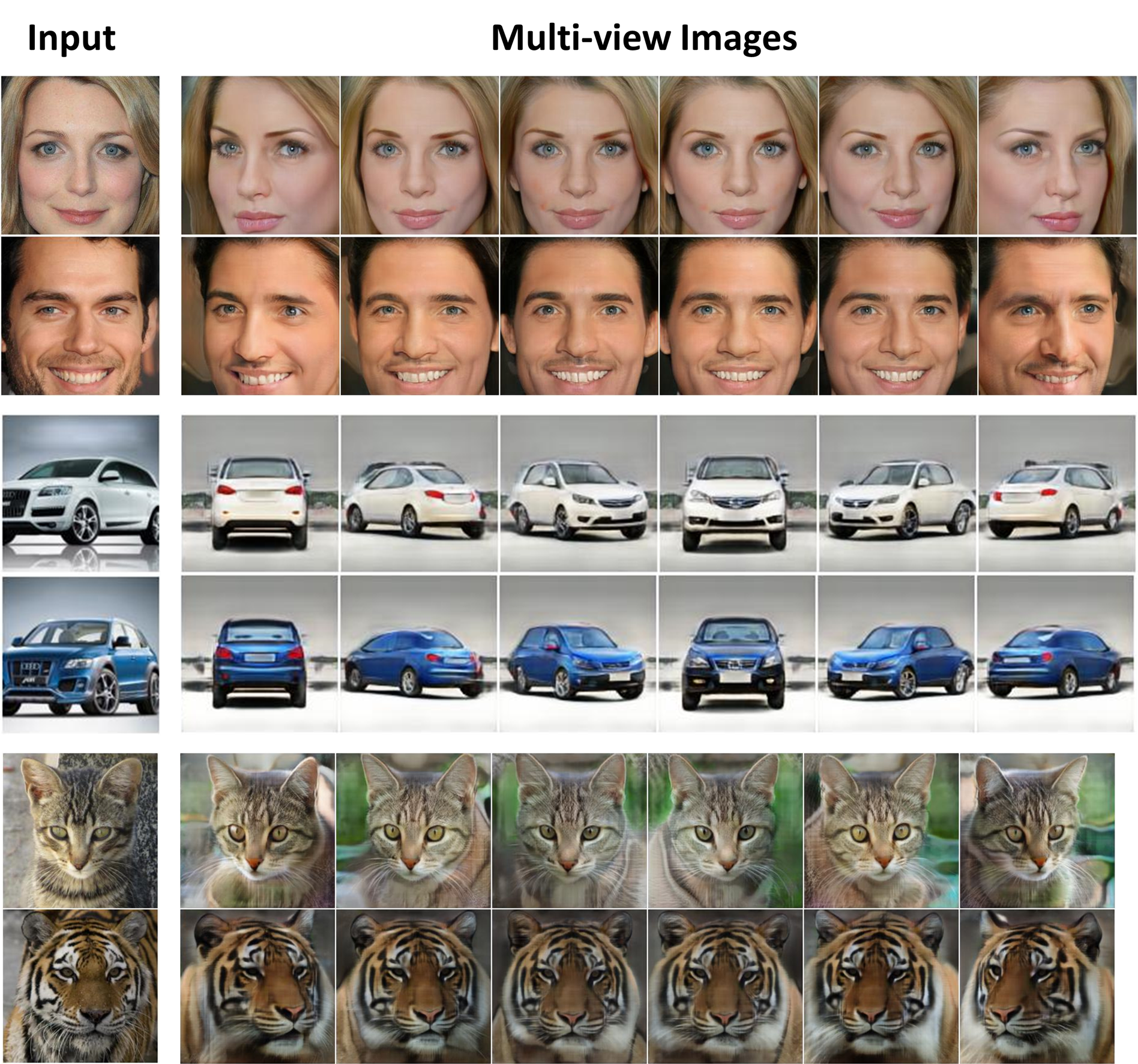}}
    \caption{{\bf Demonstration of the 3D reconstruction results employing our proposed method, ZIGNeRF.} This illustration depicts the successful zero-shot 3D GAN inversion across various real-world image datasets \cite{choi2020stargan,karras2017progressive, yang2015large}.}
    \label{figure:figure_1}
\end{figure}

\section{Introduction}
The remarkable success of generative adversarial networks (GANs) \cite{goodfellow2020generative} has spurred significant advancements in realistic image generation with high quality. Particularly, following the emergence of StyleGAN \cite{karras2019style}, numerous 2D-based generative adversarial network models have benefited from a deeper understanding of latent spaces \cite{karras2021alias,karras2020analyzing}. Consequently, various computer vision tasks, such as conditional image generation and style transfer \cite{hollein2022stylemesh, ko2023superstargan}, have shown substantial progress. However, 2D-based image generation models are constrained in their ability to generate novel view images due to their limited understanding of the underlying 3D geometry of real-world scenes.

To overcome this challenge, several studies have adopted the neural radiance field (NeRF) \cite{mildenhall2021nerf} approach, which encodes a scene into a multi-layer perceptron (MLP) to provide 3D rendering. Although conventional NeRF \cite{mildenhall2021nerf} has successfully facilitated the development of 3D-aware models and reduced computational costs in novel view synthesis tasks, it remains impractical to train a model overfitted to a single scene with multi-view images \cite{mildenhall2021nerf, yu2021pixelnerf}. Consequently, various studies have extended NeRF by integrating it with generative models, i.e., generative NeRF. Generative NeRF \cite{chan2022efficient, chan2021pi, deng2022gram, gu2021stylenerf, niemeyer2021giraffe, or2022stylesdf, schwarz2020graf} models can be trained on unposed real-world images, whereas conventional NeRF necessitates multiple images of a single scene \cite{sitzmann2019deepvoxels, tancik2022block, xu2022point}. Moreover, generative NeRF has been employed for obtaining conditional samples through techniques such as class label information \cite{jo2021cg} or text encoding \cite{gal2022stylegan, radford2021learning, ramesh2022hierarchical, wei2022hairclip}.

Despite the convenience and intuitiveness of these approaches, they possess limitations in image editing and generating 3D representations of specific inputs, such as out-of-domain images or real-world images. To enable more practical applications, generative NeRF models have also incorporated GAN inversion techniques \cite{richardson2021encoding, roich2022pivotal, song2022editing, zhu2020domain} for the 3D representation of particular input images, including out-of-distribution or real-world images. However, previous studies have faced a constraint that necessitates fine-tuning on pre-trained models for specific images \cite{ko20233d, lin20223d, xie2023high, yin2023nerfinvertor}. This requirement hinders the application of these models to numerous real samples simultaneously and renders the process time-inefficient, as it demands extensive fine-tuning.

In this study, we propose a novel zero-shot methodology for the generation of multi-view images, derived from input images unseen during the training process. This approach leverages a 3D-aware GAN inversion technique. Notably, our model proffers 3D-consistent renderings of unposed real images during inference, eliminating the need for supplementary fine-tuning.

Our architectural design bifurcates into two distinct components: the 3D-generation module and the 3D-aware GAN inversion module. The former is founded on the principles of GIRAFFE \cite{niemeyer2021giraffe}, which successfully amalgamates the compositional attributes of 3D real-world scenes into a generative framework. To enhance the precision of 3D real-world reconstruction and improve image quality, we introduce modifications to the GIRAFFE module, specifically in the decoder and neural renderer. The 3D-aware GAN inverter, on the other hand, is trained with images synthesized from the generator. This strategic approach enables the inverter to accurately map the input image onto the generator's manifold, regardless of the objects' pose. Example results of our model is displayed in Fig. \ref{figure:figure_1}.

We subject our model to rigorous evaluation, utilizing five diverse datasets: Cats, CelebA, CelebA-HQ, AFHQ, and CompCars. Additionally, we demonstrate the model's robustness by inputting FFHQ images into a model trained on CelebA-HQ.
The primary contributions of this work are as follows:
\begin{itemize}
    \item We present ZIGNeRF, a pioneering approach that delivers a 3D-consistent representation of real-world images via zero-shot estimation of latent codes. To our knowledge, this is the first instance of such an approach in the field.
    \item ZIGNeRF exhibits robust 3D feature extraction capabilities and remarkable controllability with respect to input images. Our model can perform 3D operations, such as a full 360-degree rotation of real-world car images, a feat not fully achieved by many existing generative NeRF models.
\end{itemize}

\section{Related Work}
\subsection{Neural Radiance Field (NeRF)}
NeRF is an influential method for synthesizing photorealistic 3D scenes from 2D images. It represents a 3D scene as a continuous function using a multi-layer perceptron (MLP) that maps spatial coordinates to RGB and density values, and then generates novel view images through conventional volume rendering techniques. Consequently, NeRF significantly reduces computational costs compared to existing voxel-based 3D scene representation models \cite{henzler2019escaping, nguyen2019hologan, seitz1999photorealistic, sitzmann2019deepvoxels, zhou2018stereo}. However, the training method of NeRF, which overfits a single model to a single scene, considerably restricts its applicability and necessitates multiple structured training images, including camera viewpoints \cite{chang2015shapenet, sitzmann2019deepvoxels}.

\subsection{Generative NeRF}
Generative NeRFs optimize networks to learn the mapping from latent code to 3D scene representation, given a set of unposed 2D image collections rather than using multi-view supervised images with ground truth camera poses. Early attempts, such as GRAF \cite{schwarz2020graf} and pi-GAN \cite{chan2021pi}, demonstrated promising results and established the foundation for further research in the generative NeRF domain. Recent works on generative NeRF have concentrated on generating high-resolution 3D-consistent images. The recently proposed StyleNeRF \cite{gu2021stylenerf} successfully generates high-resolution images by integrating NeRF into a style-based generator, while EG3D \cite{chan2022efficient} exhibits impressive results with a hybrid architecture that improves computational efficiency and image quality.

However, real-life applications frequently necessitate conditional samples that exhibit the desired attribute rather than random samples in data distribution. We adopt GAN inversion as a conditional method, as opposed to class-based or text encoding conditional methods, which are prevalent in 2D generative models \cite{choi2018stargan}. The aforementioned conditional generation techniques, such as class-based or text encoding methods, possess limitations. Firstly, the training dataset must include conditional information, such as labels or text corresponding to each sample. Secondly, they cannot provide 3D representation of real-world images as conditional input. We address these limitations in existing conditional generative NeRF models by introducing GAN Inversion into generative NeRF for conditional generation.

\subsection{3D aware GAN inversion}
With the remarkable progress of GANs, numerous studies have endeavoured to understand and explore their latent space to manipulate the latent code meaningfully. GAN inversion represents the inverse process of the generator in GANs. Its primary objective is to obtain the latent code by mapping a given image to the generator's latent space. Ideally, the latent code optimized with GAN inversion can accurately reconstruct an image generated from the pre-trained generator. The output sample can be manipulated by exploring meaningful directions in the latent space \cite{shen2020interpreting}. Moreover, real-world images can be manipulated in the latent space using GAN inversion.

Several studies have investigated 3D GAN inversion with generative NeRF to generate multi-view images of input samples and edit the samples in 3D manifolds. Most previous works fine-tuned the pre-trained generator due to the utilization of optimization-based GAN inversion methods. However, additional steps for fine-tuning the generator for GAN inversion impose limitations in terms of adaptability and computational costs.

In this paper, we propose a novel inverter for 3D-aware zero-shot GAN inversion. The proposed inverter can map out-of-domain images into the latent space of the generator. Our model can generate 3D representations of real-world images without requiring additional training steps. The proposed 3D-aware zero-shot GAN inversion maximizes applicability since the trained model can be directly applied to out-of-domain images.

\begin{figure}[t]
    \centerline{\includegraphics[width=\columnwidth]{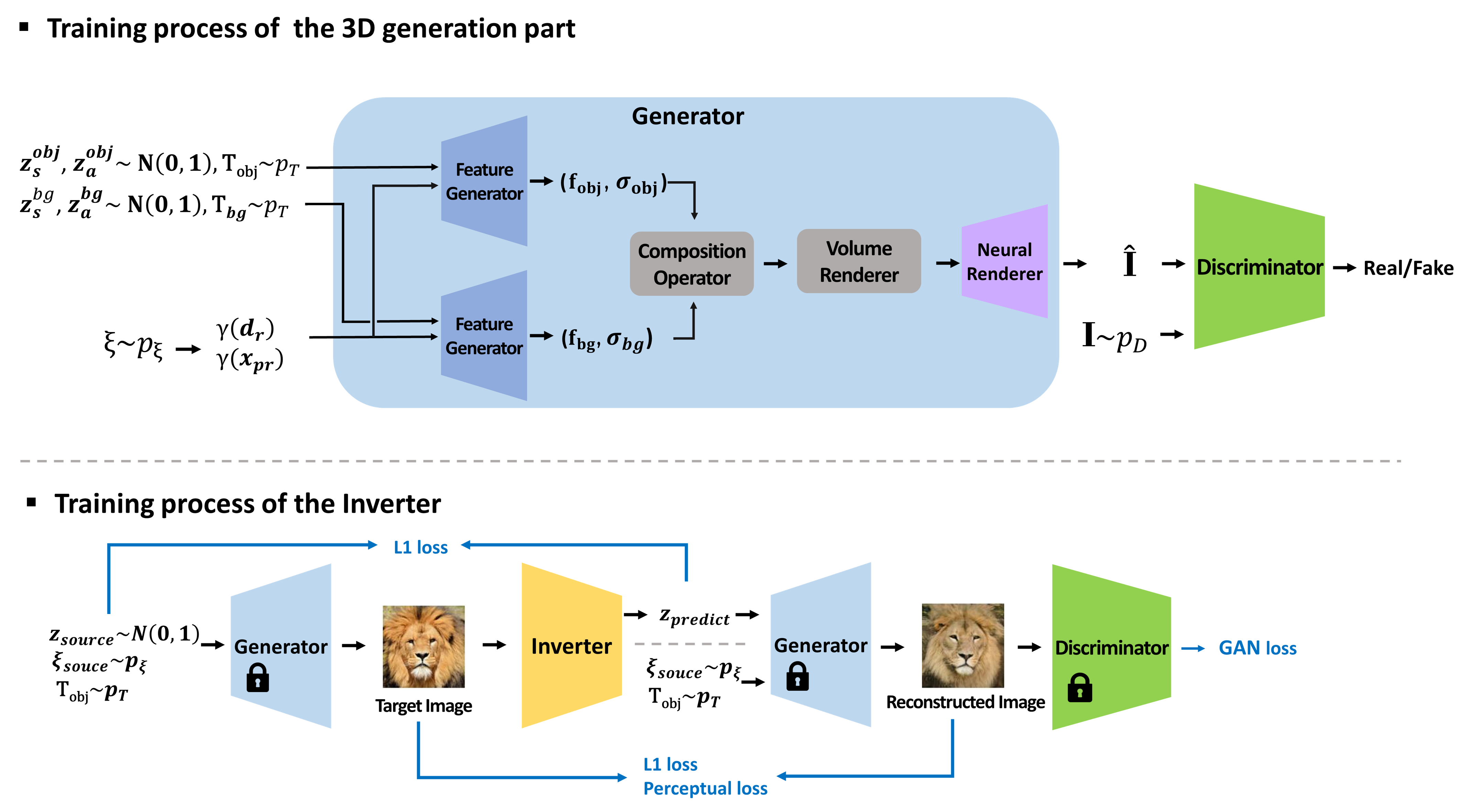}}
    \caption{{\bf The comprehensive architecture of ZIGNeRF.} The 3D generative component is trained to produce photorealistic images consistent with 3D structures by mapping the latent code and camera pose to a synthetic image. Subsequently, the inverter is trained in conjunction with the pre-trained generator and discriminator.}
    \label{figure:figure_2}
\end{figure}

\section{Method}
This work seeks to generate multi-view images from an out-of-domain image by combining generative NeRF with GAN inversion. The proposed method, graphically delineated in Fig. \ref{figure:figure_2}, encompasses two distinct phases: the 3D-generation segment and the 3D-aware inverter. The first phase involves training the 3D-generation component, an architecture based on GIRAFFE, augmented by enhancements in the neural renderer and the discriminator modules to fortify and expedite the training process. In the second phase, the 3D-aware inverter is trained with the pre-trained generator. The novel inverter is designed to transform out-of-domain images into latent codes within the generator's latent space. Consequently, the generator can produce multi-view images of the out-of-domain image using the latent code derived from the inverter. Throughout the training of the inverter, we utilize the images generated from the generator, imbued with 3D information, as the training dataset. At test time, the inverter executes zero-shot inversion on real-world images, obviating the need for additional fine-tuning for unseen images. The proposed method thereby holds great promise for generating 3D-consistent multi-view images from real-world input images.

\subsection{3D Generation}
{\bf Compositional Generative Neural Feature Field.}\; Our 3D-generator represents a scene with a compositional generative neural feature field, a continuous function inherited from GIRAFFE, to represent a scene. This is essentially a combination of feature fields, each representing an object in a single scene, with the background also considered an object. In the 3D-generator, a 3D location, ${\mathbf x} \in\ \mathbb{R}^3$, a viewing direction, ${\bf d } \in\ \mathbb{S}^2$, and latent code, ${\bf z} \sim {\mathcal{N}(0,1)}$, are mapped to a volume density $\sigma \in\ \mathbb{R}^+$ and a high-dimensional feature field ${\bf {f }} \in\ \mathbb{R}^{M_f}$, rather than RGB colour ${\bf{c }} \in\ \mathbb{R}^3$. 

Affine transformation is applied to objects in the scene so that each object can be controlled in terms of poses, which include scale, translation, and rotation:

\begin{equation}
    T =  \left\{\textbf{s}, \textbf{t}, \textbf{R}\right\},
\end{equation}

where \textbf{s}, \textbf{t} $\in\ \mathbb{S}$ indicate scale and translation parameters, respectively, and \textbf{R} $\in$ SO(3) determine rotation. The affine transformation enables object-level control by generating the bounding box corresponding to T of a single object:

\begin{equation}
   \tau = \textbf{R}\cdot \textbf{s}\textbf{I}\cdot +\textbf{t},
\end{equation}

where \textbf{I} is the 3 × 3 identity matrix. Compositional generative neural feature field is parameterized with an MLP as follows:

\begin{equation}
   C((\sigma_i, {\mathbf{f}}_i)^{N}_{i=1})=C(f_{\theta i}(\gamma(\tau^{-1}({\textbf{x}})), \gamma(\tau^{-1}({\textbf{d}})), {\textbf{z}}_i)^{N}_{i=1}),
\end{equation}
\begin{equation}
   z = [{\bf{z}^1_s}, {\bf{z}^1_a}, ..., {\bf{z}^N_s}, {\bf{z}^N_a}],
\end{equation}

where $\gamma \left( \cdot \right)$ is positional encoding function \cite{mildenhall2021nerf}, which is applied separately to x and d, and $C \left( \cdot \right)$ is the compositional operator that composites feature field from the N-1 objects and a background. We then volume render the composited volume density and feature field rather than directly output the final image. 2D-feature map, which is fed into neural renderer for final synthesized output, is attained by volume rendering function $\pi_v$, 

\begin{equation}
    \pi_{v}(C(\sigma, {\textbf{f}}))={\textbf{F}}.
\end{equation}

{\bf Neural renderer with residual networks.}\; Our model outputs final synthetic image with neural rendering on the output feature map of volume rendering. We observe that the original neural renderer of GIRAFFE does not preserve the feature well. Furthermore, the learning rate of the decoder and the neural renderer is not synchronized; hence the training of the generator is unstable.\

We improve the simple and unstable neural renderer of GIRAFFE. Our neural renderer replaces 3×3 convolution layer blocks with residual blocks \cite{he2016deep} and employs the ReLU activation rather than leaky ReLU activation \cite{xu2015empirical} for faster and more effective rendering. To stabilize the neural rendering, we adopt spectral normalization \cite{miyato2018spectral} as weight normalization. We experimentally verify that the modified neural renderer improves the stability of the training and the quality of the outputs. Our neural renderer, which maps the feature map F to the final image $\hat{I} \in\ \mathbb{R}^{H\times W\times 3}$, is parameterized as:

\begin{equation}
    \pi_{\theta}(\bf{F})=\hat{I}.
\end{equation}

{\bf Discriminator.}\; As the vanilla GAN \cite{goodfellow2020generative}, the discriminator outputs probability, which indicates whether the input image is real or fake. We replace the 2D CNN-based discriminator with residual blocks employing spectral normalization as weight normalization.

{\bf Objectives.}\; The overall objective function of the 3D-generative part is:
\begin{equation}
    L_{\text{{G, D}}}={L_{\text{GAN}}}+\lambda L_{\text{R1}},
\end{equation}

where $\lambda$ = 10. We use GAN objective \cite{goodfellow2020generative} with R1 gradient penalty \cite{mescheder2018training} to optimize the network.

\subsection{3D-aware Invertor}
\begin{figure}[h]
    \centerline{\includegraphics[width=0.3\columnwidth]{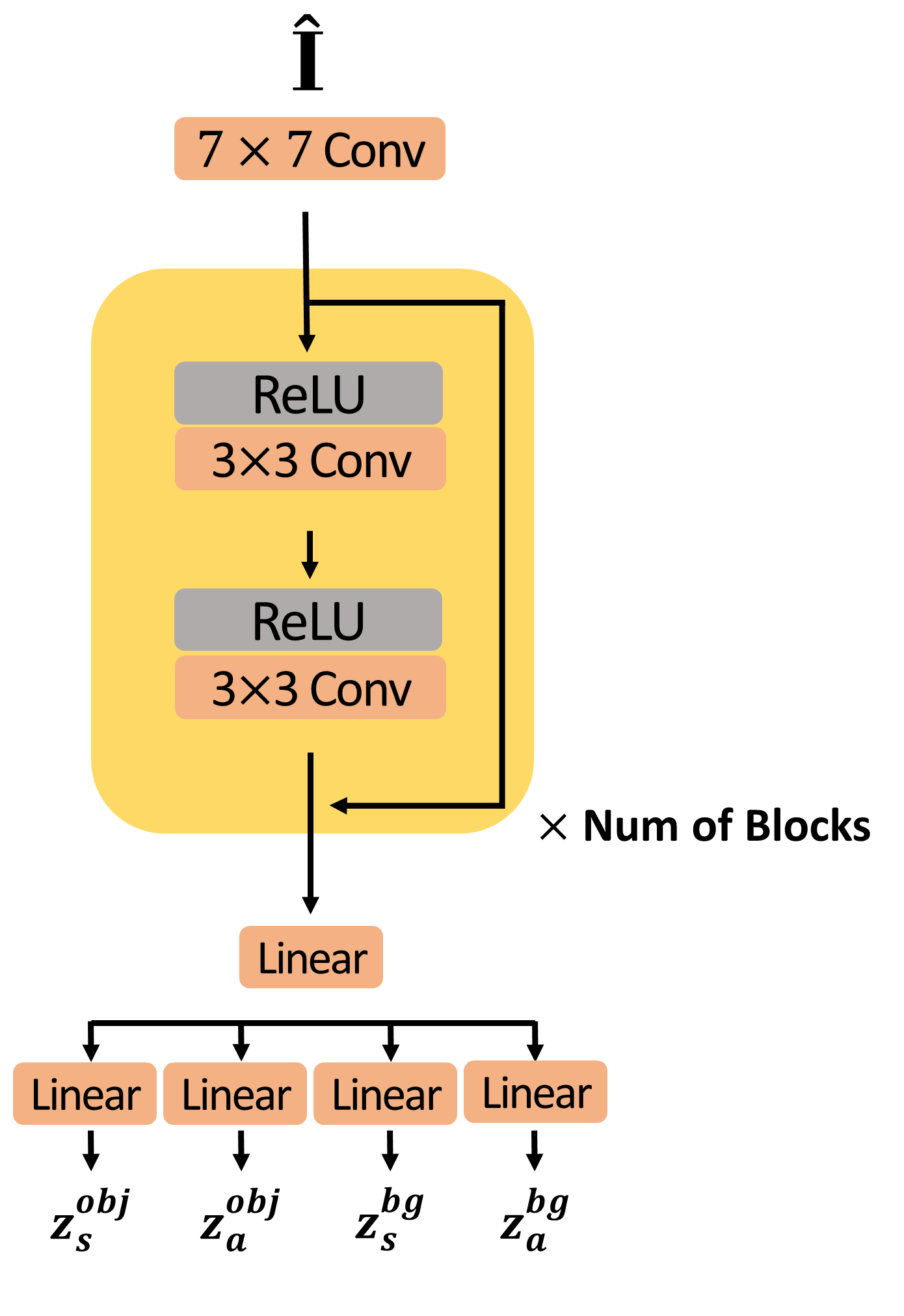}}
    \caption{{\bf Schematic representation of the architecture of the inverter deployed in ZIGNeRF.}}
    \label{figure:figure_3}
\end{figure}

To invert a given image into latent codes within the generator's latent space, we introduce a novel inverter. This inverter is designed by stacking the residual encoder block with ReLU activations, as depicted in Fig. \ref{figure:figure_3}. Four linear output layers are situated at the culmination of the inverter to facilitate output. These residual blocks extract the feature of the input image, and each linear output layer estimates the {$\bf{z}^{obj}_s$}, {$\bf{z}^{obj}_a$}, {$\bf{z}^{bg}_s$}, {$\bf{z}^{bg}_a$} of the input image.

The challenge of 3D-aware GAN inversion involves mapping multi-view images of a single object into a unique latent code. To construct a 3D-aware inverter, we opt to use the synthesized image $\hat{\textbf{I}}$ as the training data. Given that we already possess the source parameters of the generated image, the inverter solely estimates the latent code {$\bf{z}^{predict}$} of the input image. The generated training images equip the inverter to extract the feature of unseen images, which vary in viewing direction, scale, and rotation. Following the latent code inference, the pre-trained generator reconstructs the input image using {$\bf{z}^{predict}$} and source parameters, which include camera pose, $\boldsymbol{\xi}^{\bf{source}}$, and compositional parameter, ${\bf{T}^{source}} = \left\{\textbf{s}, \textbf{t}, \textbf{R}\right\}$:

\begin{equation}
    I_{\theta}(\hat{\bf{I}}) = {\bf{z}^{predict}},
\end{equation}

\begin{equation}
    G_{\theta}(\bf{z}^{predict}, \bf{T}^{source} ,\boldsymbol{\xi}^{source})= \bf{\hat{I}}^{reconst}.
\end{equation}

As the inverter learns to estimate the latent source code, we found that the L1 loss between the two latent codes in latent space was inadequate for reconstructing the scene. Thus, we opted to employ GAN loss and L1 as an image-level loss to generate a plausible image. In addition, we incorporated two perceptual losses, namely the Structural Similarity Index Measure (SSIM) \cite{wang2004image} and the Learned Perceptual Image Patch (LPIPS) \cite{zhang2018unreasonable} loss, to conserve the fine details of the source image. The inverter can be optimized using the following function:

\begin{align*}
    L_I & =L_{\text{GAN}}({\bf{\hat{I}^{predict}}})
    +\lambda_1 L_{\text{latent}}({\bf{z^{source}}}, {\bf{z^{predict}}})\\
    &+\lambda_2 L_{\text{reconst}}({\bf{\hat{I}^{source}}}, {\bf{\hat{I}^{predict}}})
    +\lambda_3 L_{\text{percept}}({\bf{\hat{I}^{source}}}, {\bf{\hat{I}^{predict}}}),
\end{align*}

where  $\bf{\hat{I}}^{predict}$ indicates the image reconstructed by the pre-trained generator using $\bf{z}^{predict}$. $L_\text{latent}$ and $L_\text{reconst}$ represent latent-level and image-level loss, respectively, both utilizing L1 loss. $L_\text{percept}$ signifies image-level perceptual loss, employing the LPIPS loss and SSIM loss.

\subsection{Training specifications}

During the training phase, we randomly sample the latent codes $\bf z_s, \bf z_a \sim {\mathcal{N} (0,1)}$, and a camera pose $\boldsymbol{\xi} \sim {p_{\xi}}$. The parameters $\lambda_1$, $\lambda_2$, and $\lambda_3$ are set to 10, 100, and 1, respectively, for training the inverter. The model is optimized using the RMSProp optimizer \cite{ruder2016overview}, with learning rates of 1 × $10^4$, 7 × $10^5$, and 1 × $10^4$ for the generator, the discriminator, and the inverter, respectively. We utilize a batch size of 32. For the first 100,000 iterations, the generator and the discriminator are trained, and the inverter is trained for the next 50,000 iterations. During the training process of the inverter, the generator and the discriminator remain frozen.

\begin{figure}[t]
    \centerline{\includegraphics[width=\columnwidth]{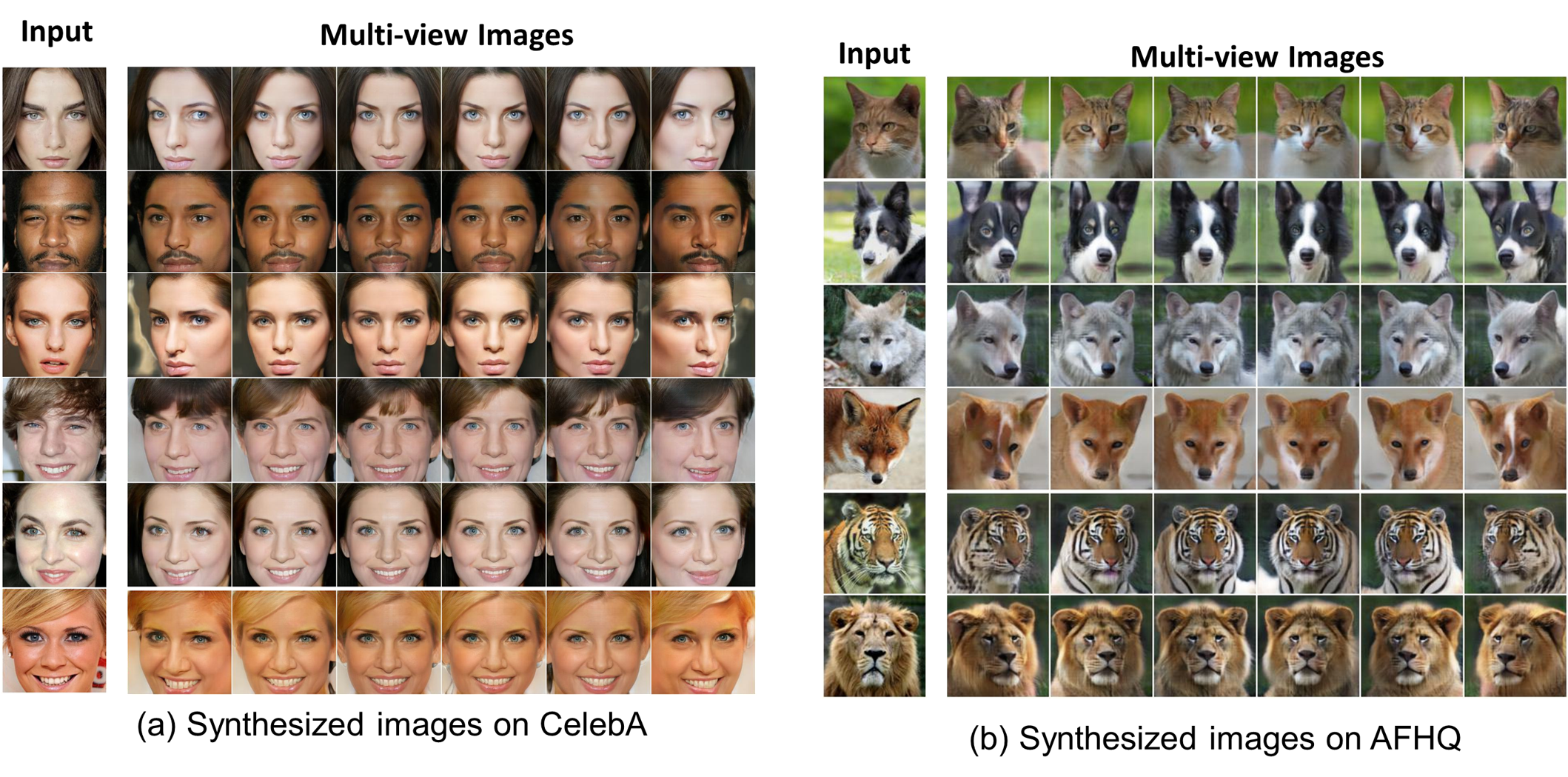}}
    \caption{{\bf Display of $\bf{256}^2$ multi-view synthesis applied to facial datasets: CelebA-HQ \cite{karras2017progressive} and AFHQ \cite{choi2020stargan}.}}
    \label{figure:figure_4}
\end{figure}

\begin{figure}[t]
    \centerline{\includegraphics[width=\columnwidth]{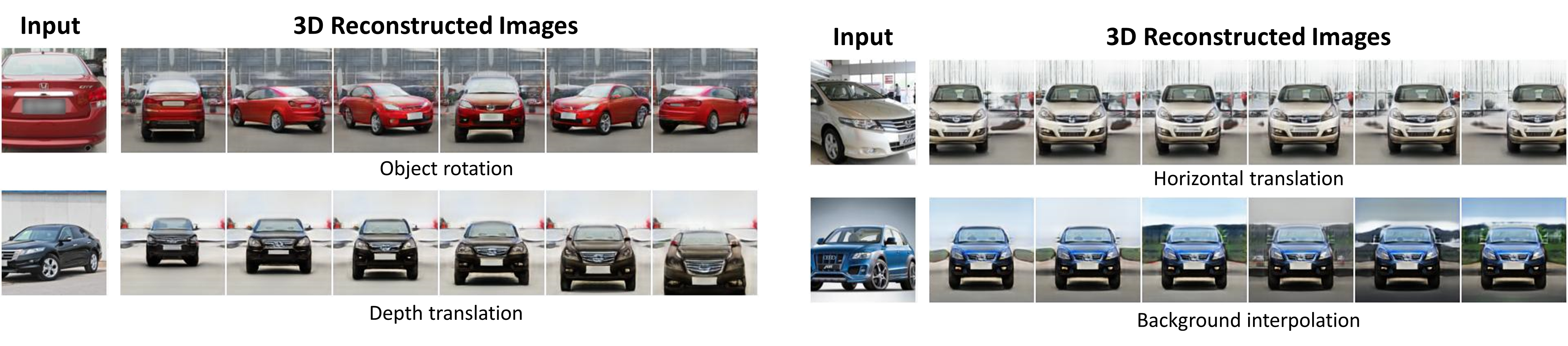}}
    \caption{{\bf Visualisation of reconstructed images based on an input car image \cite{yang2015large}, following compositional operations.} These illustrations highlight the effective disentanglement of the object from the background and the provision of 3D controllability.}
    \label{figure:figure_5}
\end{figure}

\section{Experiments}
ZIGNeRF is evaluated concerning zero-shot feature extraction, 3D controllability, and adaptability. We test on five real-world datasets: Cats, AFHQ \cite{choi2020stargan}, CelebA \cite{liu2015deep}, CelebA-HQ \cite{karras2017progressive}, and CompCar \cite{yang2015large}. An additional dataset, FFHQ \cite{karras2019style}, is used to demonstrate the robust adaptation capabilities of the proposed model. All input images shown in this section were not used during the training process, thereby validating the zero-shot 3D GAN inversion with unseen images. We commence with a visual validation of the proposed model, examining the similarity between the input image and the reconstructed images and 3D-consistent controllability. The model is then evaluated using Fréchet Inception Distance (FID) \cite{heusel2017gans} as a metric. We conclude with ablation studies to validate the efficacy of the loss function in optimizing the inverter.

\subsection{Controllable 3D Scene Reconstruction} 
\begin{figure}[t]
    \centerline{{\includegraphics[width=0.7\columnwidth]{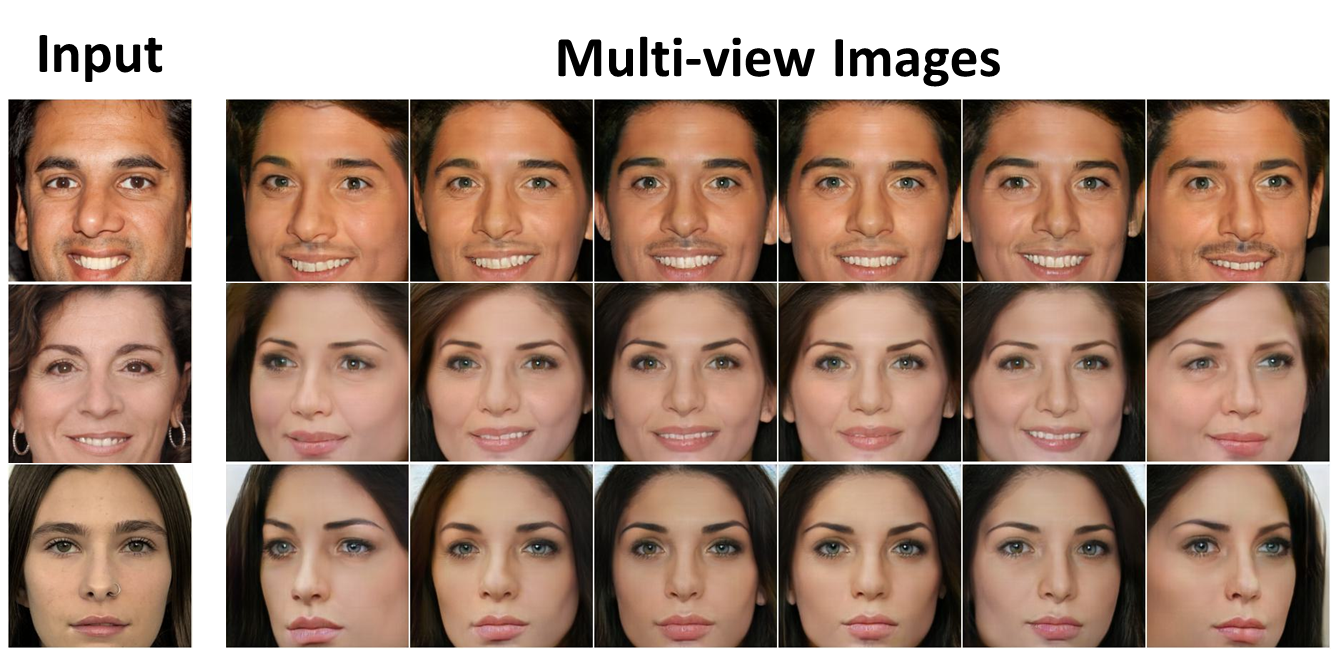}}}
    \caption{{\bf Presentation of $\textbf{256}^2$ synthesized images conditioned on input FFHQ \cite{karras2019style} images, produced by the model trained on the CelebA-HQ dataset \cite{karras2017progressive}.}}
    \label{figure:figure_6}
\end{figure}

We visually demonstrate that our proposed model generates multi-view consistent images corresponding to the input image. Fig. \ref{figure:figure_4} showcases 3D reconstruction on CelebA-HQ \cite{karras2017progressive} and AFHQ \cite{choi2020stargan}, substantiating that the inverter successfully extracts facial features irrespective of gender or skin colour in human faces, and species in animal faces. Fig. \ref{figure:figure_5} exhibits the model's controllability and object disentanglement with CompCar \cite{yang2015large}, indicating that the inverter estimates the latent code of the object and background effectively. Notably, the proposed model can facilitate 3D-consistent 360-degree rotation, a common limitation of generative NeRF methods. We further attest to the robustness of our model by applying it to FFHQ, as shown in Fig. \ref{figure:figure_6}.

\begin{table}[t]
\renewcommand{\arraystretch}{1.5}
\vspace{-0.1cm}
\centering
\resizebox{\columnwidth}{!}{%
\begin{tabular}{cccccccccc}
\toprule[1.5pt]
\multirow{2}{*}{Method} & 
\multirow{2}{*}{Models} & 
    \multicolumn{2}{c}{Cats} &
    \multicolumn{2}{c}{CelebA(HQ)} &
    \multicolumn{2}{c}{CompCar} &
    \multicolumn{2}{c}{AFHQ} \\
    &
    &
    $128^2$ &
    $256^2$ &
    $128^2$ &
    $256^2$ &
    $128^2$ &
    $256^2$ &
    $128^2$ &
    $256^2$ \\
    \hline\hline
\multirow{2}{*}{Unconditional} &
    GIRAFFE &
    24.01 &
    21.28 &
    19.45 &
    23.14 &
    38.91 &
    40.84 &
    35.03 &
    38.18 \\
    &
    ZIGNeRF(ours) &
    \textbf{12.31} &
    \textbf{11.21} &
    \textbf{11.01} &
    \textbf{14.98} &
    \textbf{22.67} &
    \textbf{22.57} &
    \textbf{12.81} &
    \textbf{19.96} \\
    \hline
    Conditional &
    ZIGNeRF(ours) &
    \textbf{15.06} &
    \textbf{16.83} &
    \textbf{14.77} &
    \textbf{25.66} &
    \textbf{25.97} &
    \textbf{25.41} &
    \textbf{14.02} &
    \textbf{28.78} \\
    \bottomrule[1.5pt]
\end{tabular}%
}
\vspace{0.2cm}
\caption{{\bf Comparative analysis of the FID between our proposed ZIGNeRF and a baseline model.} The models were trained on four distinct datasets with the resolution of $128^2$ and $256^2$.}
\label{Table 1.}
\end{table}

\subsection{Quantitative Evaluation} 
To thoroughly evaluate the efficacy of our proposed model, ZIGNeRF, we conduct experiments in both conditional and unconditional generation modes. The evaluation process involves a random sampling of 20,000 real images alongside 20,000 synthesized images, which is a conventional method to compare generative models. The results are displayed in Tab. 1.

In the context of the unconditional model, we generate samples using random latent codes. The training process entails 100,000 iterations. Notably, our model, ZIGNeRF, significantly outperforms the baseline GIRAFFE \cite{niemeyer2021giraffe} model. As an illustration, for the CelebA(HQ) $256^2$ dataset, ZIGNeRF achieves a score of 14.98, substantially lower than the GIRAFFE's score of 23.14. This is indicative of the model's ability to produce higher-quality images with fewer iterations.

Turning to the conditional synthesis, the latent codes estimated by the inverter are employed on randomly sampled real images. The training process for the generator is conducted over 100,000 iterations, while the inverter training comprises 50,000 iterations, during which the generator is kept static. When compared to GIRAFFE, ZIGNeRF demonstrate superior performance in conditional samples as well. For instance, in the AFHQ $128^2$ dataset, our model attains a score of 14.02, marking a significant improvement over the GIRAFFE's score of 35.03.

\begin{figure}[h]
    \centerline{{\includegraphics[width=0.4\columnwidth]{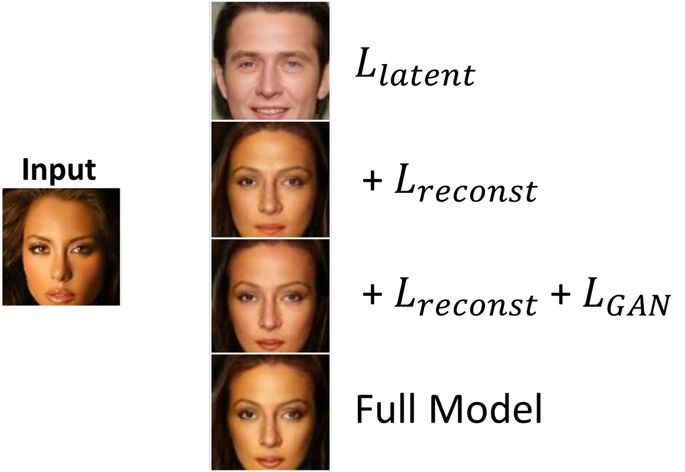}}}
    \caption{{\bf \bf Ablation study of the loss functions employed in the training of the inverter within ZIGNeRF.}}
    \label{figure:figure_7}
\end{figure}

\subsection{Ablation study}
In the interest of validating the loss function deployed in training the inverter, we undertake an ablation study. The study scrutinizes the necessity of each loss component: latent loss, reconstruction loss, GAN loss, and perceptual loss. The imperative nature of each loss function is demonstrated through its incremental addition to the naive model, which is trained solely via latent code comparison. Fig. 7 illustrates the individual contribution of each loss function. It is observed that the naive model exhibits limited capability in reconstructing the input image. The reconstruction loss $L_\text{reconst}$ aligns the reconstructed image with the input at an image-level. The GAN loss $L_\text{GAN}$ is observed to enhance the realism of the reconstructed image, independent of improving the input-reconstructed image similarity. The full model elucidates that the perceptual loss $L_\text{percept}$  plays a pivotal role in refining the expression of minute attributes, skin colour, and texture.

\section{Conclusion}
In this paper, we have proposed ZIGNeRF, an innovative technique that manifests a 3D representation of real-world images by infusing a 3D-aware zero-shot GAN inversion into generative NeRF. Our inverter is meticulously designed to map an input image onto a latent manifold, a learning process undertaken by the generator. During testing, our model generates a 3D reconstructed scene from a 2D real-world image, employing a latent code ascertained from the inverter. Rigorous experiments conducted with four distinct datasets substantiate that the inverter adeptly extracts features of input images with varying poses, thereby verifying the 3D controllability and immediate adaptation capabilities of our model.

Our novel approach carries the potential for wide application, given that our pipeline can be generally applied to other existing generative NeRFs. It is worth noting that this zero-shot approach is a pioneering contribution to the field, bringing forth a paradigm shift in 3D image representation. In future work, we envisage extending the proposed method by manipulating the inverted latent code for editing the input image, thereby further enhancing the capabilities of this innovative model.

\newpage

\bibliographystyle{abbrv}  
\bibliography{references}

\newpage

\appendix

\renewcommand{\thefigure}{S\arabic{figure}}
\renewcommand{\thetable}{S\arabic{table}}

\section{Introduction of Supplementary Material}
In this supplemental document, we offer a detailed overview of the various architectural elements within the network – including the feature fields, the neural renderer, and the discriminator, all discussed in Section \ref{sec:secA}. Furthermore, in Section \ref{sec:secB}, we elucidate a quantitative analysis of our ablation study results, underscoring the efficacy of our loss functions during the training phase of the inverter. In conclusion, we bring forth additional qualitative findings on datasets such as CelebA-HQ \cite{karras2017progressive}, CompCar \cite{yang2015large}, and AFHQ \cite{choi2020stargan}. Two novel experimental approaches are also introduced: the style-mixed 3D representation of two facial input images, and the generation of two objects within a single scene using a generator trained on single-object scenes.


\begin{figure}[h]
    \centerline{\includegraphics[width=0.7\columnwidth]{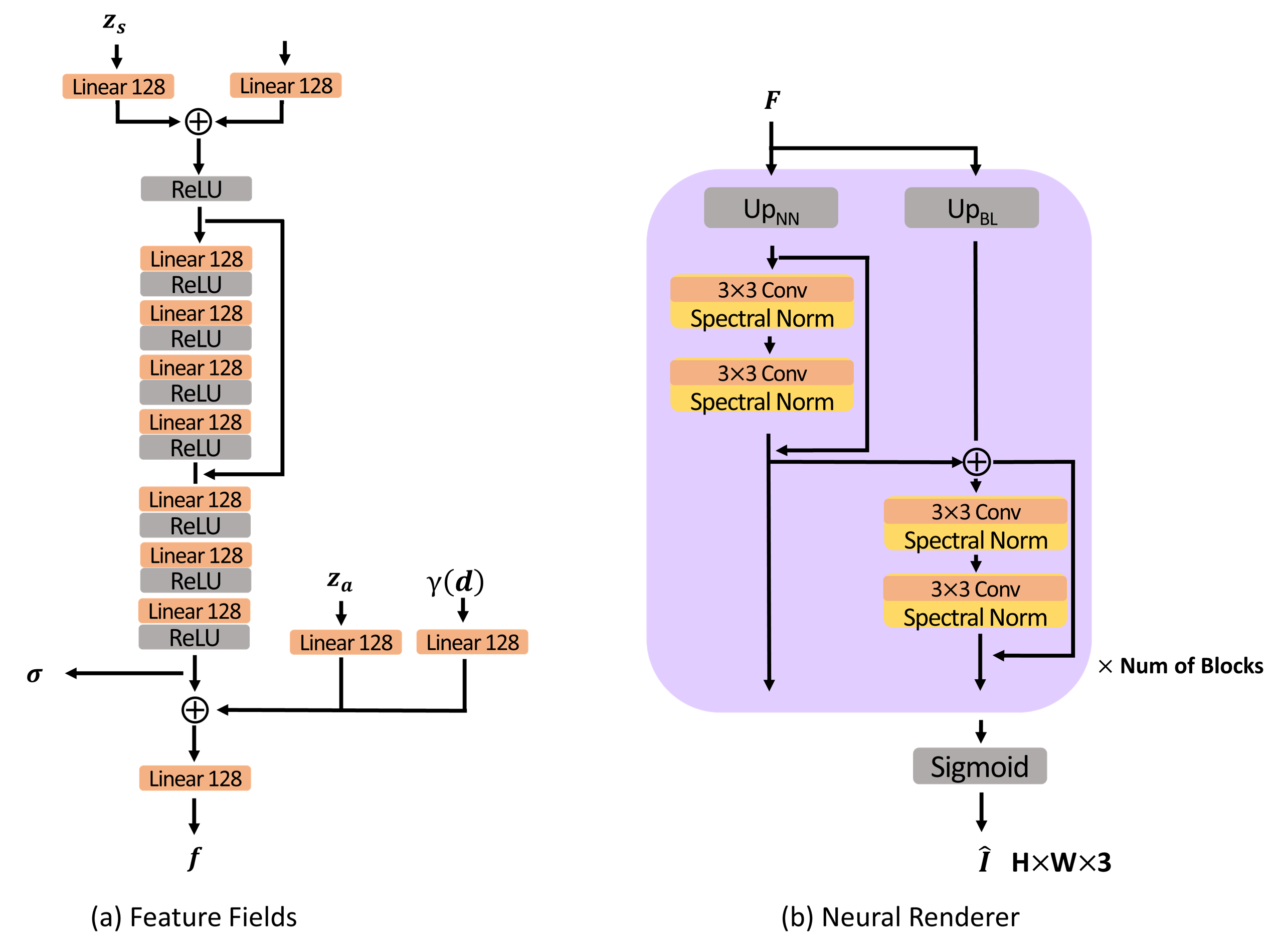}}
    \caption{{\bf Architecture of the feature fields and neural renderer.} The feature fields are parameterized with multi-layer perceptrons (MLPs) as shown in the (a). The 3D point {\bf x}, viewing direction {\bf d}, and latent codes ${\bf z_s}$, ${\bf z_a}$ are mapped into a volume density $\sigma$ and feature \textbf{f}. In (b), the neural renderer blocks depict the transformation of the volume-rendered feature image F into final synthesized image $\hat{I}$ . $UP_{NN}$ and $UP_{BL}$ symbolize the nearest neighbour upsampling and bilinear upsampling, respectively.}
    \label{figure:figure_S1}
\end{figure}

\section{Network Architectures} \label{sec:secA}
\begin{figure}[t]
    \centerline{\includegraphics[width=0.3\columnwidth]{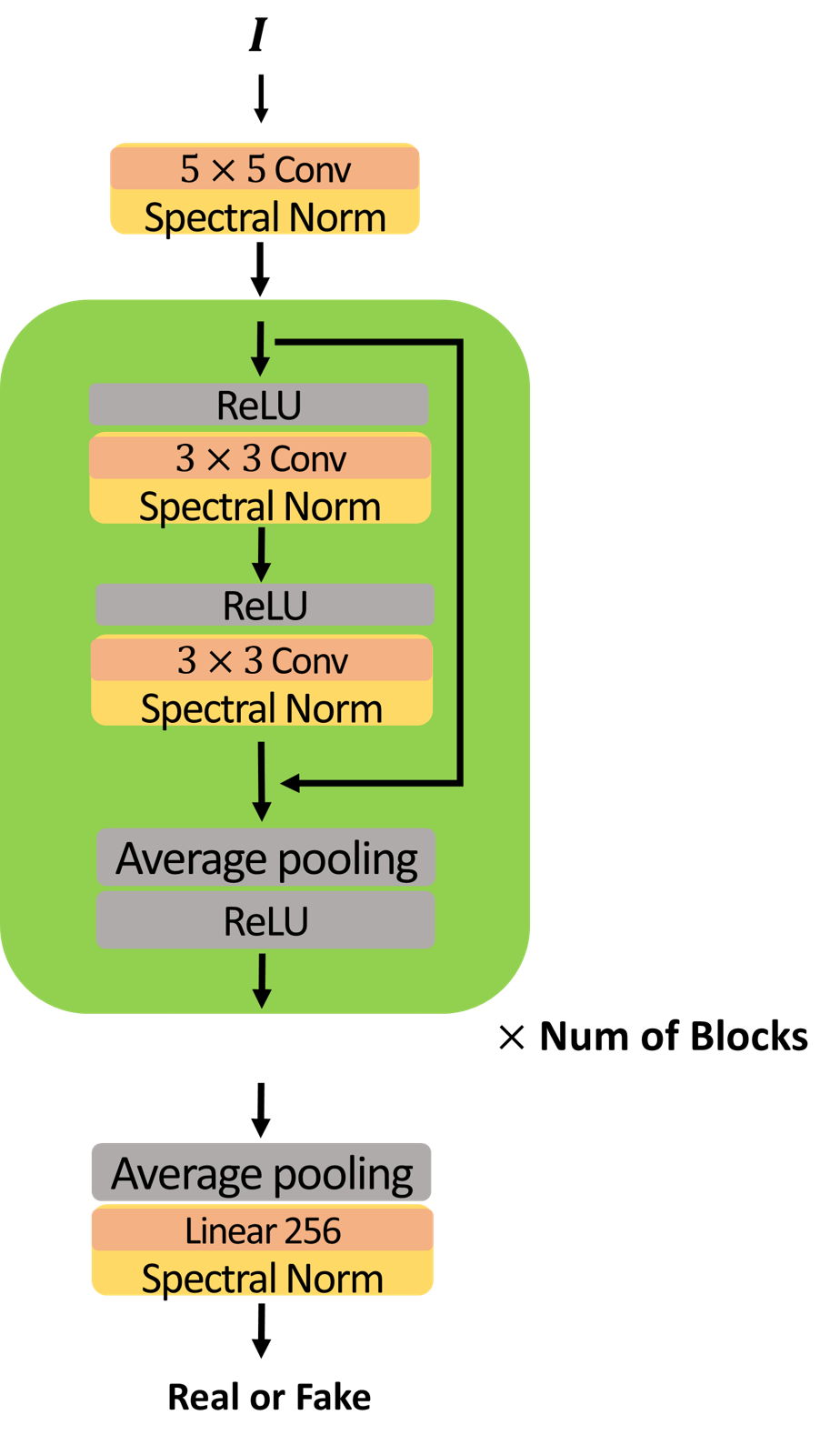}}
    \caption{{\bf Architecture of the discriminator.} The input image is processed through residual convolution blocks fortified with spectral normalization, and an average pooling operation. The output probability is derived after the final linear layer with spectral normalization.}
    \label{figure:figure_S2}
\end{figure}

In this section, we provide the details of network architecture: feature fields, neural renderer, and the discriminator as exhibited in Fig. \ref{figure:figure_S1} and \ref{figure:figure_S2}.

Fig. \ref{figure:figure_S1} presents a detailed overview of the architecture underpinning the feature fields and the neural renderer. The construct of the feature fields is parameterized via multi-layer perceptrons, colloquially referred to as MLPs, a feature vividly displayed in subfigure (a). This setup maps a three-dimensional point, the viewing direction, along with latent codes into a volume density and a feature. Subfigure (b) unravels the process behind the neural renderer blocks, demonstrating how these blocks transform a volume-rendered feature image, into the ultimate synthesized image. 

Fig. \ref{figure:figure_S2} explicates the architecture of the discriminator network, emphasizing the steps involved in processing the input image. Initially, the image is subjected to a series of residual convolution blocks, which are fortified with spectral normalization. This is followed by the execution of an average pooling operation. The process culminates with the derivation of the output probability, which is obtained post the final linear layer, again, involving spectral normalization.

\begin{table}[h]
\renewcommand{\arraystretch}{1.5}
\vspace{-0.1cm}
\centering
\resizebox{0.25\columnwidth}{!}{%
\begin{tabular}{lc}
\toprule[1.5pt]
Ablation Losses & FID \\
\hline \hline
$L_{latent}$ & 80.08 \\
+$L_{reconst}$ & 17.82 \\
+$L_{GAN}$ & 15.53 \\
\hline
\textbf{Full model} & 14.77 \\
\bottomrule[1.5pt]
\end{tabular}%
}
\vspace{0.2cm}
\caption{{\bf FID score of the ablation study.} The full model has been trained with latent loss, reconstruction loss, GAN loss, and perceptual loss.}
\label{Table:Tabel_s2.}
\end{table}

\section{Supplementary Experimental Results} \label{sec:secB}
\subsection{The Necessity of Loss Components in Training Session: A Quantitative Evaluation}
Tab. \ref{Table:Tabel_s2.} offers a quantitative testament to the indispensable nature of the loss components used in the training session of the inverter. These encompass latent loss, reconstruction loss, GAN loss, and perceptual loss. It is observed that the Fréchet Inception Distance (FID) \cite{heusel2017gans} experiences a steady enhancement with each loss component incrementally added to the naive model, which originally only employs the latent loss.

\subsection{Extended Operational Results}
\begin{figure}[h]
    \centerline{\includegraphics[width=0.3\columnwidth]{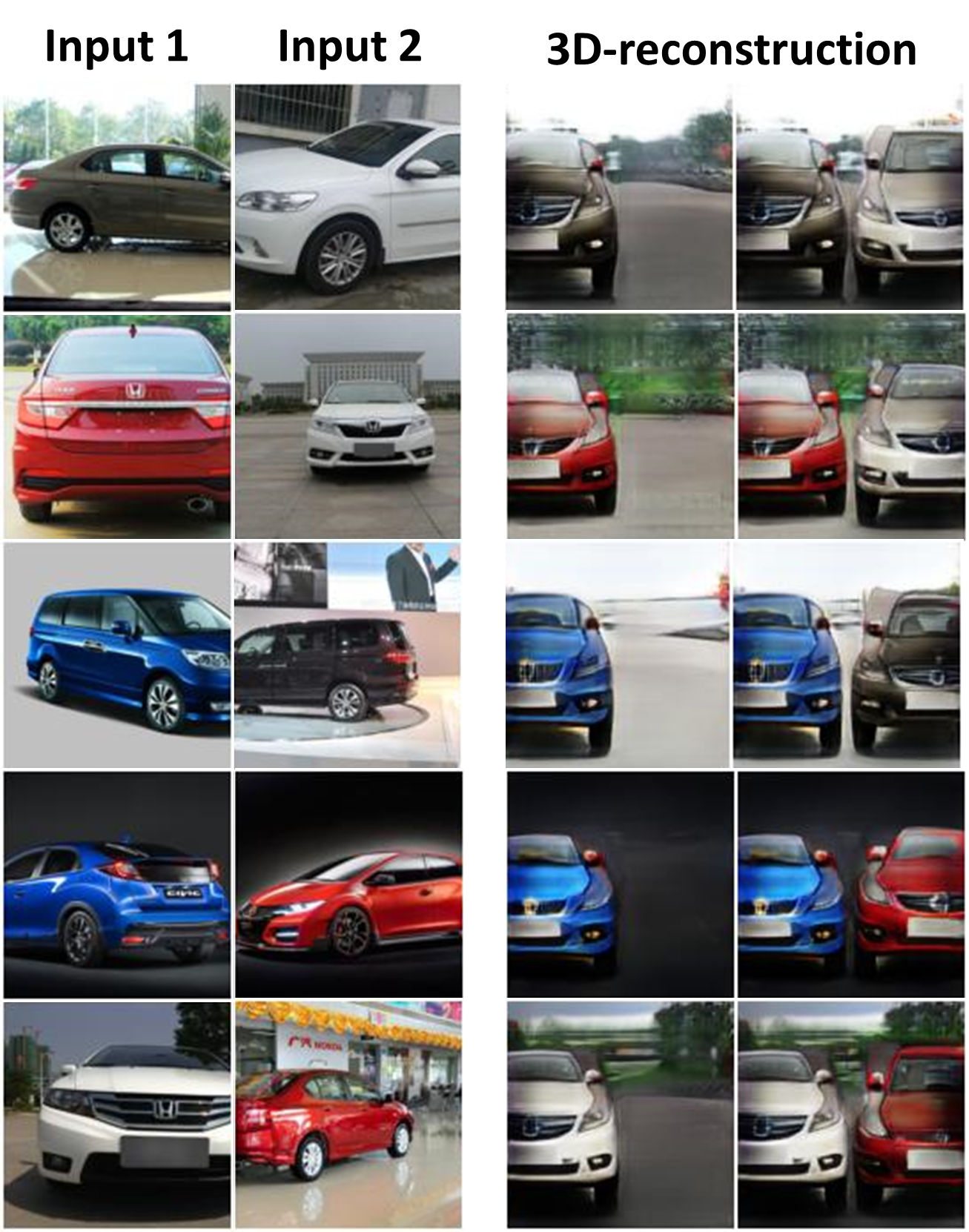}}
    \caption{{\bf Generating two objects in a single scene.} Results exhibit the compositional scene representation by generating two objects in a single scene. The invertor transforms two input images into two sets of the latent codes, and the generator which trained on single-object scenes synthesizes a single scene including two independent objects.}
    \label{figure:figure_S3}
\end{figure}

\begin{figure}[h]
    \centerline{\includegraphics[width=0.7\columnwidth]{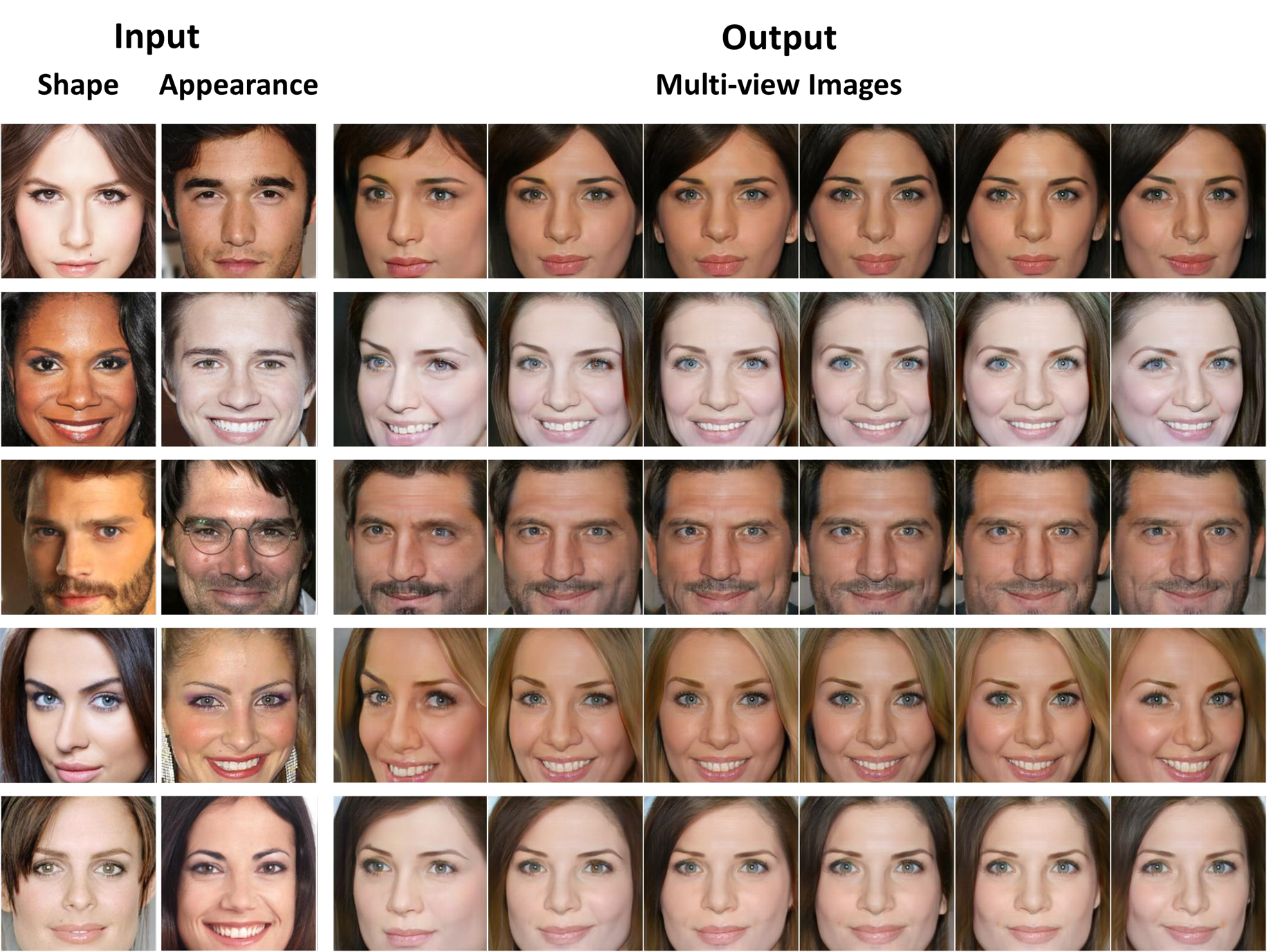}}
    \caption{{\bf Multi-view images with style mixing of two CelebA-HQ input images.} The invertor extracts the latent codes from two independent input images for generating style mixed images. Each output object is generated by $\bf z_s$  of the first image and $\bf z_a$ of the second image.}
    \label{figure:figure_S4}
\end{figure}

In this section, we present the application results of the proposed model through Fig. \ref{figure:figure_S3} and \ref{figure:figure_S4}, showcasing style-mixed 3D synthesis and the generation of two objects within a single scene.

Our model demonstrates a unique ability to generate multiple objects within a single scene, even when trained on a dataset consisting primarily of single-object scenes. This is accomplished by leveraging multiple decoder segments within our network architecture. Although our empirical exploration has only been executed on one dataset, the theoretical underpinnings suggest a promising generalizability of this phenomenon. A testament to the robustness of our model is its successful exhibition of zero-shot learning capabilities, as evidenced by an experiment where two CompCars images are synthesized into one image. Like the generation of individual objects, each object within the composite scene retains the ability to undergo transformations such as longitudinal displacement and rotation.

Additionally, we incorporate style mixing in our model with the application of the inverter structure we proposed, utilizing the CelebA-HQ dataset. In the style mixing paradigm that we suggest, our inverter, producing two distinct outputs, generates a shape vector from one image, and an appearance vector from another. These two vectors are subsequently utilized as input for the generator to synthesize a novel object. This process further underscores the model's zero-shot learning capability.

\subsection{Supplementary Results}
\begin{figure}[h]
    \centerline{\includegraphics[width=0.8\columnwidth]{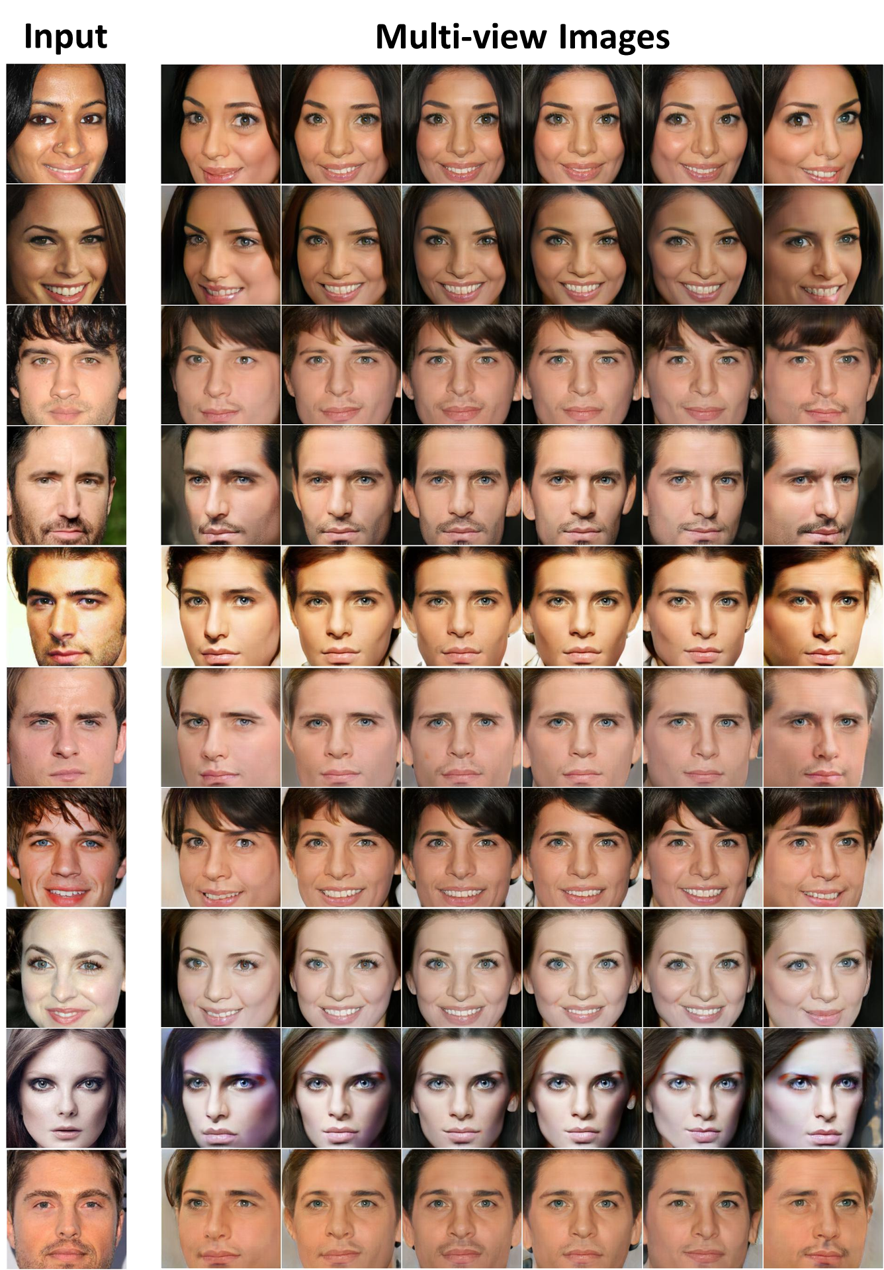}}
    \caption{{\bf Supplementary results with $\bf{256}^2$ CelebA-HQ image inputs.}}
    \label{figure:figure_S5}
\end{figure}

\begin{figure}[t]
    \centerline{\includegraphics[width=\columnwidth]{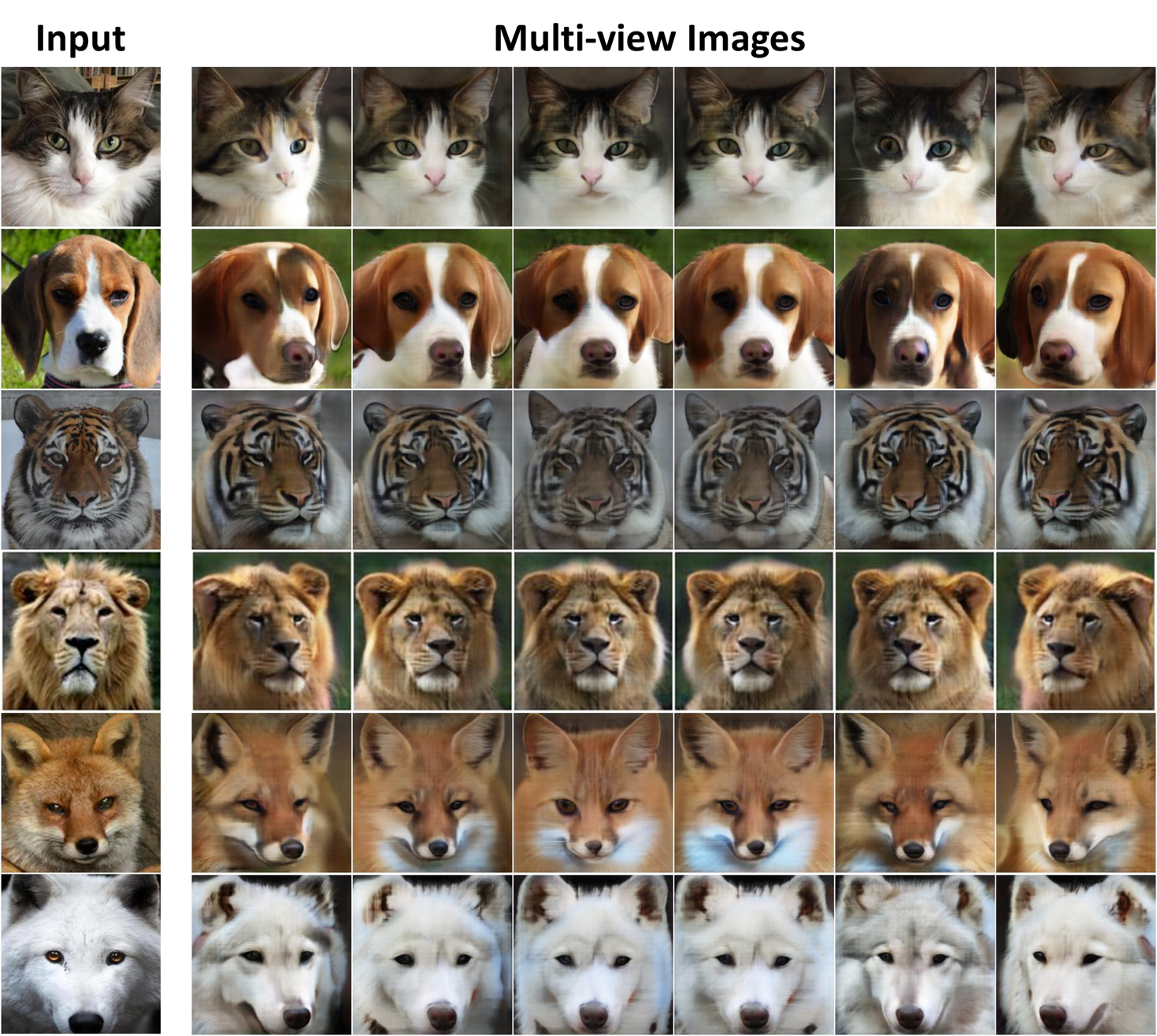}}
    \caption{{\bf Supplementary results with $\bf{256}^2$ AFHQ image inputs.}}
    \label{figure:figure_S6}
\end{figure}

\begin{figure}[t]
    \centerline{\includegraphics[width=\columnwidth]{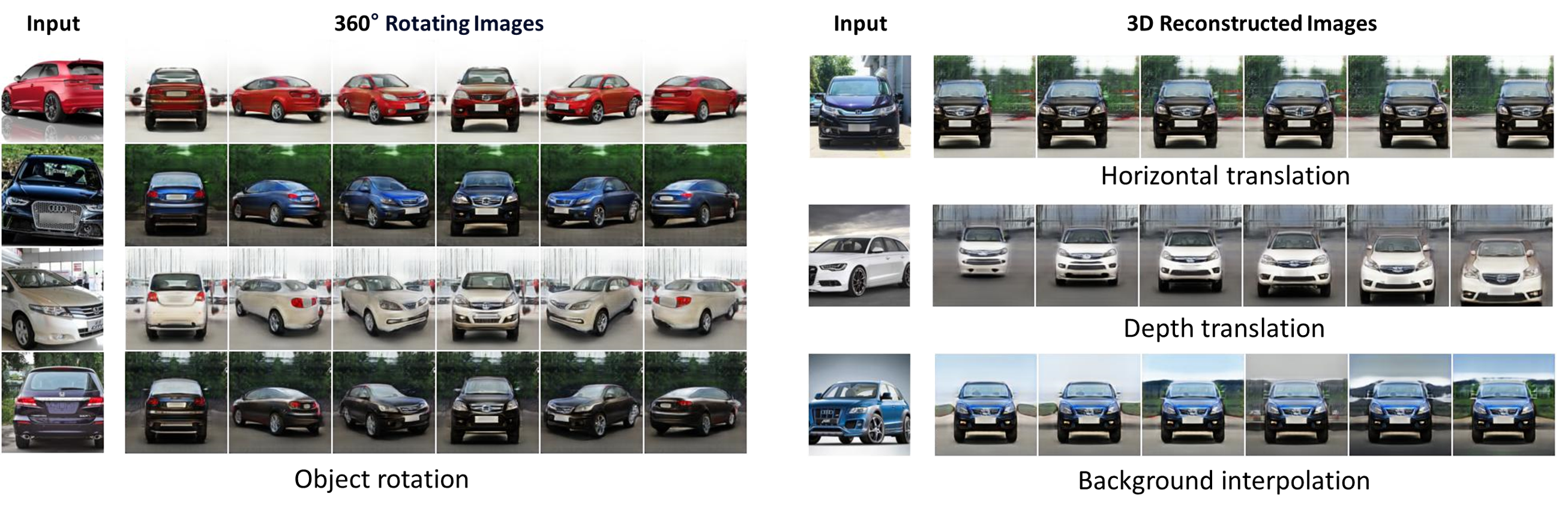}}
    \caption{{\bf Controllable image synthesis with $\bf{128}^2$ CompCars image inputs.}}
    \label{figure:figure_S7}
\end{figure}

Fig. \ref{figure:figure_S5}, \ref{figure:figure_S6}, and \ref{figure:figure_S7} deliver additional examples on CelebA-HQ \cite{karras2017progressive}, AFHQ \cite{choi2020stargan}, and CompCar \cite{yang2015large} datasets. 

We embark on rigorous evaluation of our model using a diverse range of input images sourced from varied datasets. With the CelebA dataset, we assess the model's performance using faces of different genders, ages, and ethnic backgrounds, all of which yield impressive quality in output.
In the context of the AFHQ dataset, we utilize images from a variety of categories as input for our testing phase. It is worth noting that these results, encompassing distinct categories, are obtained using a single model with different conditional vector inputs, thereby highlighting the large capacity of our model.

The CompCars dataset allows us to experiment with 360-degree image generation using real image inputs representing various car models, colours, and camera poses. It is important to note that a significant advantage of our model is the freedom it provides in the longitudinal movement of objects, along with the capacity to alter the background. This flexibility underpins the model's capacity for highly controllable image synthesis, an attribute that holds immense potential for a wide array of applications.

\end{document}